\documentclass[journal]{IEEEtran}
\ifCLASSINFOpdf
\else
\fi

\usepackage{graphics} 
\usepackage{epsfig} 
\usepackage{mathptmx} 
\usepackage{times} 
\usepackage{amsmath} 
\usepackage{amssymb}  
\usepackage{bm}
\usepackage{mathrsfs}
\usepackage{xcolor}
\usepackage{cite}
\usepackage{threeparttable}
\usepackage{multirow}
\usepackage{bigdelim}
\usepackage{algorithm}
\usepackage{algorithmicx}
\usepackage{algpseudocode}
\usepackage{graphicx}
\usepackage{subfigure}
\usepackage{comment}

\begin{document}
%
\title{Cellular Decomposition for Non-repetitive Coverage Task with Minimum Discontinuities}
%
%
%


\author{Tong Yang$^1$, Jaime Valls Miro$^2$~\IEEEmembership{Member,~IEEE,}, Qianen Lai$^1$, Yue Wang$^{1*}$ and Rong Xiong$^1$
\thanks{$^1$ Tong Yang, Qianen Lai, Yue Wang and Rong Xiong are with the State Key 
Laboratory of Industrial Control and Technology, Zhejiang University, P.R. China. 
}
\thanks{$^2$ Jaime Valls Miro is with the Centre for Autonomous Systems (CAS), University of Technology Sydney (UTS), Sydney, Australia.}
\thanks{$^*$ Corresponding Author. \newline \indent
E-mail address: {\tt\small wangyue@iipc.zju.edu.cn}}
}

%
%

\markboth{This paper is concurrently submitted for TMech and AIM 2020 Presentation}{}
%



\maketitle

\begin{abstract}
A mechanism to derive non-repetitive coverage path solutions with a proven minimal number of discontinuities is proposed in this work, with the aim to avoid unnecessary, costly end effector lift-offs for manipulators. The problem is motivated by the automatic polishing of an object. Due to the non-bijective mapping between the workspace and the joint-space, a continuous coverage path in the workspace may easily be truncated in the joint-space, incuring undesirable end effector lift-offs. Inversely, there may be multiple configuration choices to cover the same point of a coverage path through the solution of the Inverse Kinematics. 
The solution departs from the conventional local optimisation of the coverage path shape in task space, or choosing appropiate but possibly disconnected configurations, to instead explicitly explore the least number of discontinuous motions through the analysis of the structure of valid configurations in joint-space. The two novel contributions of this paper include proof 
that the least number of path discontinuities is predicated on the surrounding environment, independent from the choice of the actual coverage path; thus has a minimum. And an efficient finite cellular decomposition method to optimally divide the workspace into the minimum number of cells, each traversable without discontinuities by any arbitrary coverage path within. 
Extensive simulation examples and real-world results on a 5 DoF manipulator are presented to prove the validity of the proposed
strategy in realistic settings.
\end{abstract}


%
\IEEEpeerreviewmaketitle

\section{Introduction}
%
%
%
%

\IEEEPARstart{T}{he} non-repetitive \textit{coverage task} of a given object is an important application carried out by manipulators.
This is for instance the case of inspecting a surface for defects at close range, painting, deburring or polishing. 
The task is effectively encapsulated as the generic coverage path planning (CPP)~\cite{choset2001coverage}~\cite{galceran2013a} problem, which requires for the end-effector (EE) to traverse over all the points that define the surface of a given object exactly one time, whilst usually fulfilling additional task-specific constraints (e.g.  sustain a desired orientation of the EE with respect to the surface, maintain contact or exerting a constant EE force/torque). 
Typically, joint-space dimension is higher than the workspace's, 
and the Inverse Kinematic (IK) mapping between task and joint space is thus non-bijective.  
As a result, planning in the higher dimension joint-space cannot ensure non-repetitive visiting, 
and coverage paths are thus more suited to be designed directly in the workspace domain~\cite{Oriolo2005Motion}.  
\begin{figure}[t]
\centering
\subfigure[Relationship between joint- and work-space.]{
\includegraphics[width = 0.4\textwidth]{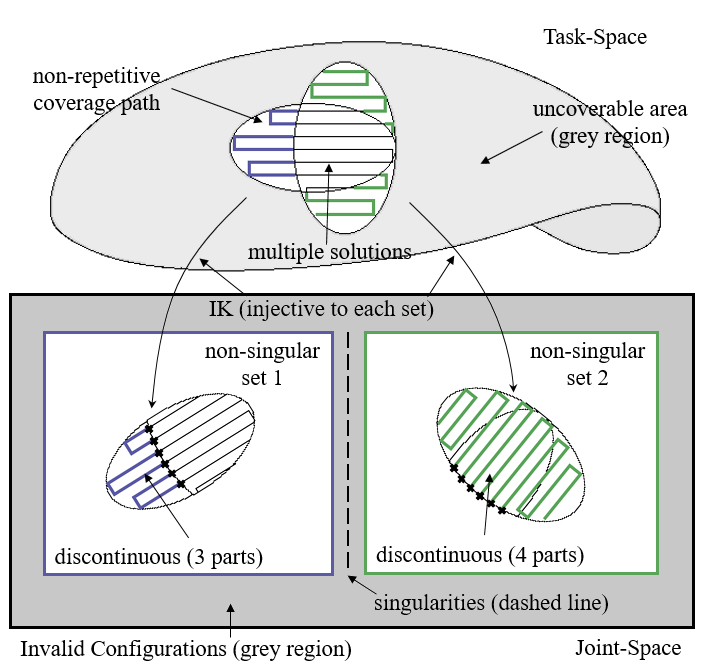}
}
\subfigure[Greedy solution example]{
\includegraphics[width = 0.22\textwidth]{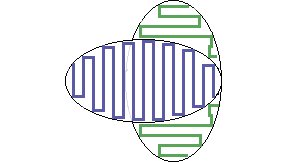}
\label{fig:greedy}
}
\subfigure[Optimal solution example]{
\includegraphics[width = 0.22\textwidth]{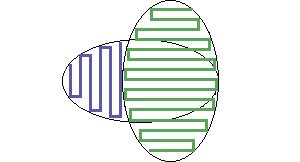}
\label{fig:optimal}
}
\caption{(a) Illustration of the coverage task problem and the relationship between joint- and work-space. 
Different colors denote disjoint sets of non-singular 
joint configurations (arbitrarily, blue may for instance represent those with 
elbow-up, whilst green may represent elbow-down), and their corresponding path in the workspace. 
The colored segments of the arbitrary coverage path shown in black have unique IK solutions. However, multiple IK solutions exist for the intersecting area shown in black.
The underlying continuous coverage path sought out thus becomes intermittent in the workspace after mapping onto the joint-space.
In this example, whatever choice among the multiple IK solutions, six discontinuities are required 
(depicted by black crosses), since the path has three separate segments in set $1$, and four in set $2$. 
The case where the joint-space solutions are taken in full from set 2 (green) is depicted.
(b) Starting from a configuration belonging to set $1$, without explicitly calculating the reachable boundary of each set, the boundary of the set $2$ within the reachable area of set $1$ is unknown. So a greedy strategy will fully cover the set $1$, dividing the uncovered region into two parts, leading to an extra lift-off. 
On the other hand, although at first sight it may appear the same as using the greedy strategy starting from set $2$ (c) illustrates the concept of CPP optimality in the joint space, whereby the continuity of the reachable area is explicitly considered, thus producing a coverage path with a single EE lift-off.} 
\label{fig1}
\end{figure}

\newpage
Yet what constitutes a continuous coverage path in the workspace may easily end up truncated into many seemingly \textit{intermittent} sections after mapping them back onto the joint-space, with undersirable path discontinuities, as graphically illustrated by Fig.~\ref{fig1}. This is also the case if a simplistic greedy strategy is followed, as the example depicted in Fig.~\ref{fig:greedy}, whereby a complete path in task space that solves for all possible configurations leads to unnecessary lift-offs to accomplish full coverage. 

Singularities have been proven to be at the origin of these bifurcations of the joint-space~\cite{porta2010path}~\cite{Porta2012Randomized}, sitting at the intersection of different configurations (e.g. elbow-up and elbow-down).
Notwithstanding singularities, for non-redundant manipulators, non-singular configurations thus form disjoint sets in the joint-space, as illustrated in Fig.~\ref{fig1}, and continuous joint-space paths between sets must then visit singularities along the way for full coverage.The problem is further compounded by additional task constraints, most notably obstacles, which produce usable configurations further divided into many disjoint sets, inevitably incurring undesirable``jumps'' between sets for successful coverage, as very rarely the whole workspace ends up mapped into a single set.  

This work advocates for the minimisation of the cost incurred on these path discontinuities, which can significantly 
outweigh any improvements proposed in the literature that may occur locally in the task space when it comes 
to coverage~\cite{hassan2018a}. This is perhaps more apparent for the case of the uniform polishing task motivating this work, 
as that means lifting the EE off the object's surface, adjusting the pose of the manipulator to the new configuration, and landing back into contact with the surface again.
This may be not only sub-optimal for the speedy completion of the CPP task, but also introduces potentially avoidable complexity in transitioning between position and force/torque control~\cite{cheah2003brief}~\cite{heck2015switched}~\cite{mirrazavi2018a}~\cite{solanes2018adaptive}~\cite{solanes2019robust} 
during the coverage task. 
 
In this work, a mechanism is proposed to address this shortcoming and derive CPP solutions with a proven minimal number of discontinuities, with the aim to avoid unnecessary, costly EE lift-offs. 
The solution departs from locally optimising the shape of the coverage path in task space, or choosing appropiate but possibly disconnected configurations, but explicitly seeking for least number of discontinuous motions through the analysis of the structure of valid configurations in the joint-space. 
The two novel contributions of this paper can be summarised as: 
\begin{enumerate}
\item Proving that the minimum number of path discontinuities, or ``lift-offs'', for the non-repetitive coverage task with non-redundant manipulators is independent of the actual choice of coverage path. 
Instead, it is predicated on the surrounding environment - the relative pose between manipulator, object and the presence of any obstacles - and this motivates to formulate the problem as a global cellular decomposition process.
On a side note, this also implies that the proposed scheme can be exploited as a criterion to evaluate the most advantageous placement of a manipulator, or object to be manipulated (e.g. polished, painted), both in a fixed configuration (automated production line), or in a mobile manipulation environment. 
\item Proposing an effective finite cellular decomposition method to divide a worskpace surface into the least number of cells whereby each is ensured to be traversable by any arbitrary inner path without incurring discontinuities.
\end{enumerate}

The remainder of this paper~\footnote{A video illustrating the concepts and results hereby described can be found here: https://www.youtube.com/watch?v=gyvXin60cCQ\&feature=youtu.be} 
is organised as follows. Section~\ref{sectionrelatedwork} reviews existing literature. 
Section~\ref{sectionproblemformulation} describes the proposed abstraction of the problem into a topological graph of surface cells corresponding 
to feasible, continuous configurations, hence administering the tools to prove that the number of path discontinuities for the CPP problem can be made independent to the eventual coverage path chosen. 
Section~\ref{sectionenumerativesolver} goes into further details about the process of finitely resolving the surface into cell elements, whilst 
Section~\ref{sectioniterativesolver} reports on the proposed iterative strategy to build on the cell elements to ensure CPP with a minimum number of discontinuitues. Experimental results from simulations and on an actual non-reduntant manipulator are collected in Section~\ref{sectionexperiment}, with final concluding remarks gathered in Section~\ref{sectionconclusion}.

\section{Related Work}\label{sectionrelatedwork}
Almost all state-of-the-art methods to solve the CPP problem first divide the robot's workspace area and then solve the CPP problem in each cell, so called cellular decomposition, which is generally further divided into two categories: exact cellular decomposition methods~\cite{lumelsky1990dynamic} and Morse-based cellular decomposition methods~\cite{choset2000exact}~\cite{Acar2002Morse}. 
Exact cellular decomposition methods divide the free space into several simple, easy sub-regions, and use conventional coverage paths, such as trapezoidal~\cite{choset2005principles} or the boustrophedon paths~\cite{choset1998coverage}~\cite{choset2000coverage}, to finish coverage in each cell. 
Morse-based cellular decomposition methods apply divisions of the free space based on the critical points of Morse functions to present more flexible shapes for cells over those extratcted by exact cellular decomposition. 
A combination of Morse decomposition and Voronoi diagrams~\cite{choset2000sensor-based} has also been proposed, particularly fitting to 
cover vast open spaces and narrow areas simultaneously.

Optimality of CPP algorithms mainly focus on metrics such as path length and time to completion.
Atkar \textit{et al.}~\cite{Atkar2003Towards} optimised the coverage path through chossing optimal starting points. 
Huang~\cite{huang2001optimal} reduced movement cost by remaining on straight paths as long as possible thus minimising the number of turns. 
Jimenez \textit{et al.}~\cite{jimenez2007optimal} used a genetic algorithm to achieve optimal coverage. 
Whilst generic, the context of these coverage works has almost invariably been motivated by mobile robots operating on 2- or 2.5-dimensional terrains. However, for manipulator, this essay advocates avoiding unnecessary path lift-off discontinuities as that decidedly outweighs any other performance metric improvement that my be achieved during the coverage process, e.g., by switching between differing geometric paths such as boustrophedon and spirals as proposed in the works of Hassan and Liu~\cite{hassan2018a}. 
These discontinuities in the CPP task are inherent to the kinematics of manipulator mechanisms, and as such the algorithms designed for mobile robots do not need to deal with this problem. 
We notice that~\cite{paus2017a} considered the pose 
optimisation of a mobile manipulator for coverage, searching for a valid criterion for the adequacy of the relative pose between manipulator and object(s) to be handled, under the assumption that repositioning the robot is costly and that simultaneous repositioning and end-effector motion is not desired. 

\section{Problem Formulation}\label{sectionproblemformulation}
In this section, we first state the problem of optimal coverage path planning ensuring least number of discontinuites.
The problem is tailored to a polishing task with the introduction of additional task-specific constraints, as per the motivation of this work. These could be waived for a more generic exercise. 
Then, we show that the least number of discontinuities is independent of the choice of physical coverage path, 
so the original problem is transformed to an optimal cellular decomposition problem. 
Finally, the minimisation problem is further transformed to a colouring problem of the derived graph ensuring least number of different colors. 

\subsection{Problem Statement}
Given the surface of an object, the kinematics of the manipulator, the shape of other obstacles in the workspace, and their relative poses, a valid coverage path consists of all valid joint-space poses of the manipulator which satisfy the following constraints:
\begin{enumerate}
\item Kinematic: the resulting manipulator motion is collision-free. When the EE contacts the surface, its $z$-axis is align with the normal vector of the surface at the contact point. 
\item Force: when the EE contacts the surface, the manipulator is able to exert the required force along the $z$-axis.  
\item Manipulability: when the EE contacts the surface, the manipulator should remain well-conditioned (under given manipulability measure~\cite{yoshikawa1990translational}), 
to dispensing with arising pertubations. 
\end{enumerate}

\noindent
The optimal CPP problem is to find a valid joint-space path whereby the manipulator EE covers the workspace non-repetitively and ensures the least number of discontinuities. A point contact between surface and EE is assumed.

\begin{figure}[t]
\centering
\includegraphics[width = 0.15\textwidth]{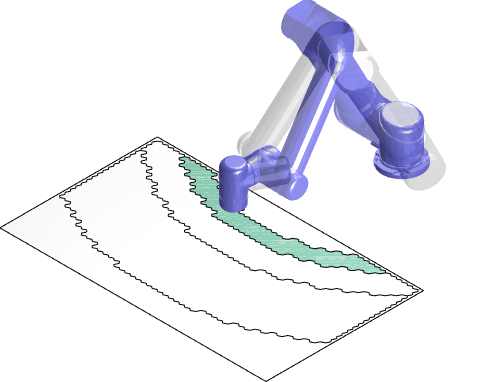}
\includegraphics[width = 0.15\textwidth]{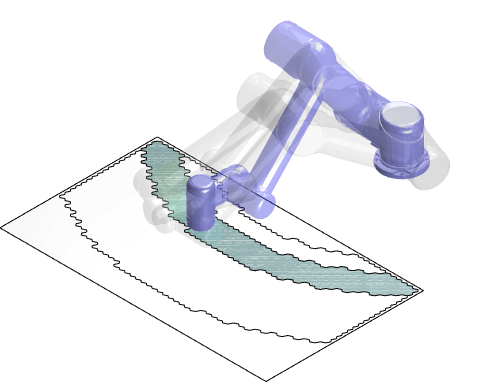}
\includegraphics[width = 0.15\textwidth]{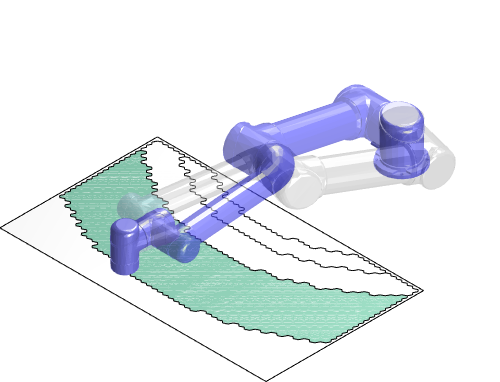}
\caption{Illustration of different IK solutions covering the same points. Vivid configurations show one class of configuration: shoulder-right, wrist-unflipped, able to reach the vivid light cyan areas. Two additional configurations (shoulder-left/right, wrist-flipped) are also valid, shown shaded, although they can only cover the middle part of the reachable area, in shaded cyan. An unpainted surface section indicates unreachable by any configuration.}
\label{figsquare}
\end{figure}

\begin{figure}[t]
\centering
\includegraphics[width = 0.44\textwidth]{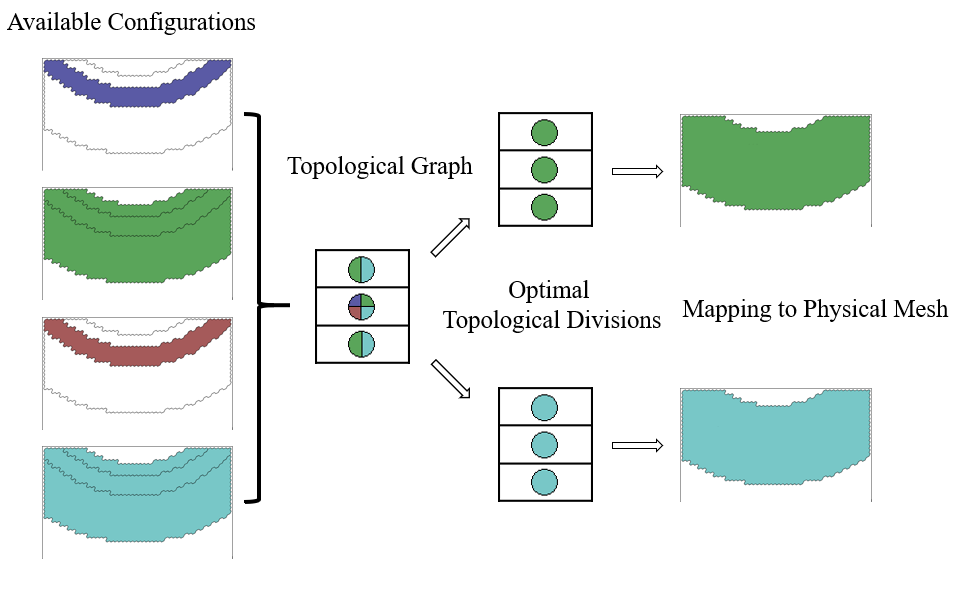}
\caption{Proposed methodology to solve the coverage problem for the set-up in Fig.~\ref{figsquare}. 
All configurations are divided into four disjoint homogeneous sets, or cells, based on their joint-space continuities, 
having the same same colour. 
The small circle filled-in with colour(s) represents all the possibilities to paint the corresponding area. A topological graph is created based on the distribution of the colors. Finally, in this example, two optimal options exist, both requiring zero lift-off.}\label{flowchart}
\end{figure}

\begin{figure}[t]
\centering
\includegraphics[width = 0.5\textwidth]{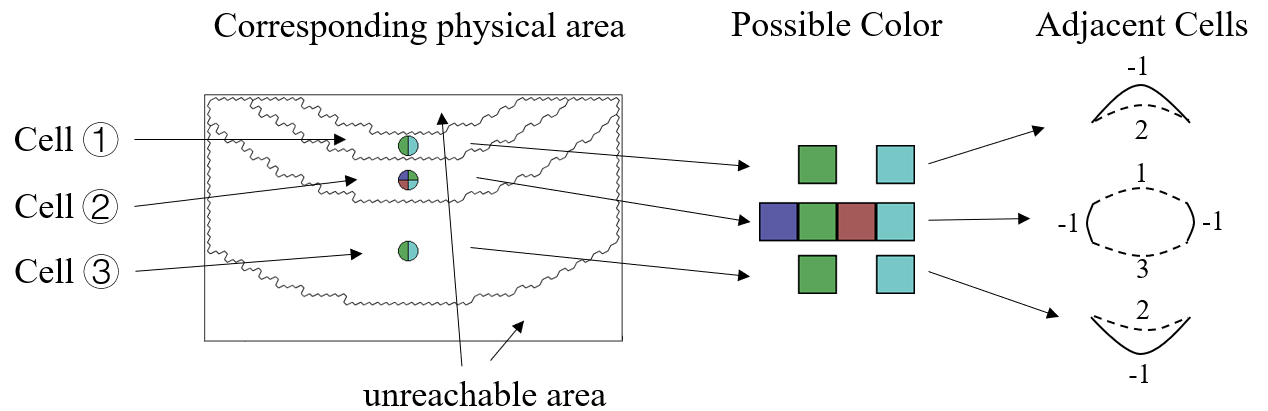}
\caption{
The elements of cells. All colors are indexed. Each cell corresponds to a connected area on the surface, and has a list of possible colors. A cell stores the index of its adjacet cell in order. 
The index of the unreachable area is denoted by $-1$. Curves are used to indicate that the edge is topological, not physical. For example, cell 2 has four possible colors (blue, green, brown and cyan) and four edges which connect to cell 1, unreachable area, cell 3 and unreachable area in order. 
}
\label{figforcolor}
\end{figure}

\subsection{Independence from the Physical Coverage Path}
An observation which simplifies the original problem is that manipulators are locally omni-directional in the joint-space, and configurations corresponding to a segment of coverage path without lift-off have high dimensional continuity in the joint-space, independently of their sequencing order. 
As a result, this work is inspired to consider only continuous regions in the joint-space and its corresponding reachable area in the workspace, instead of coverage paths in the task space, which is equivalent to a cellular decomposition problem in the workspace considering joint-space continuity. Under this equivalence, complete visiting of each cell is guaranteed with any arbitraty joint-space continuous path within, without discontinuities. 
Hence, once the cells are determined, the CPP problem within each cell 
is trivial, effectively transforming the design of the global coverage path in the traditional sense into a global cellular decomposition problem in joint-space, optimal in the number of lift-offs by incuring workspace partitions with minimum sets.

\subsection{Modeling}
Let $\mathscr{C}$ be the set of all valid configurations and $M$ be the set of all reachable points on the surface. The pose of the EE is also denoted by $M$ since there is an one-to-one correspondence between the pose of the EE and the point on the surface, so we do not distinguish them. 

 
Given a configuration $p \in \mathscr{C}$ covering $m\in M$, following the joint-space continuity, there exist a neighborhood $(p\in)U_p\subset \mathscr{C}$ that can be reached continuously (without lift-off) from $p$, covering a section of surface 
$(m\in )V_{m}\subset M$. This is illustrated in Fig.~\ref{figsquare}, where the poses reached by the manipulator 
configuration depicted in vivid colour can be reached continuously - shown in vivid cyan. 
If any of them is chosen as $p$, then all of them are in $U_p$. 
Assuming there are some other unassigned configurations, i.e., $\mathscr{C}\backslash U_p\neq \varnothing$, 
choosing another  $p'\in \mathscr{C}\backslash U_p$ specifies another set $V_{m'}\subset M$ - e.g. the shaded 
configurations depicted in Fig.~\ref{figsquare}. 
It is evident that $U_{p'} \cap U_{p} = \varnothing$. 
On repeating this process, all configurations are assigned a colour. 
Let the number of configurations be infinite. Actually, each valid configuration implies an open neighbourhood of valid configurations covering an open region on the surface (defined sub-resolutionally if the input data is discretized, like a triangular mesh in our case). The family of all open regions on the surface implied by all valid configurations is an infinite open cover of the whole surface which, with physical meanning, must have boundary. Then, even if there are infinite many open regions in the family, the Heine-Borel theorem in mathematical analysis claims the existance of a subcover with finite open regions. The finiteness of a discretized input data is trivial because the number of configurations is also finite.
As such, $\mathscr{C}$ is divided into a finite number of disjoint sets, denoted by a finite number of different colors. 

The problem also exploits the concept of a \textit{cell} defined on the task-space, following the standard terminology 
of conventional cellular decomposition methods, but with the additional property of homogeneity. 
This is established on noticing that IK mapping from reachable points in the workspace to a single set of configurations is injective, since there is no non-singular path connecting two configurations whose EEs are at a same point 
(see graphic example in Fig.~\ref{fig1}). 
The injectivity of each branch of the IK is the motivation to map the property of joint-space continuity back on to the surface, thus the algorithm can be visualized by drawing colors on the surface to form cells beloging to the same configuration class (colour).  
Refering to the same square coverage example, Fig.~\ref{flowchart} shows how $\mathscr{C}$ is divided into 4 disjoint sets. 
Since different IK solutions possess distinct colors, the available colors for points can be used to classify them. Let $\{c_i\}, i = 1, n$ be all the colors used, then for two points $m_1, m_2\in M$ their sets of available colors are $c_{m_1} = \{c_{11}, \cdots, c_{1i}\}, c_{m_2} = \{c_{21}, \cdots, c_{2j}\}$. We then say that $m_1$ and $m_2$ belong to the same cell if and only if 
$$\left\{
\begin{aligned}
& m_1,m_2 \mbox{ are connected}\\ 
& \{c_{11}, \cdots, c_{1i}\} = \{c_{21}, \cdots, c_{2j}\}
\end{aligned}
\right.
$$
Typically, for a triangular mesh surface as is our case, connectivity is provided by the edges of the mesh. Fig.~\ref{flowchart} shows the creation of the cells. 

\begin{figure}[t]
\centering
\includegraphics[width = 0.4\textwidth]{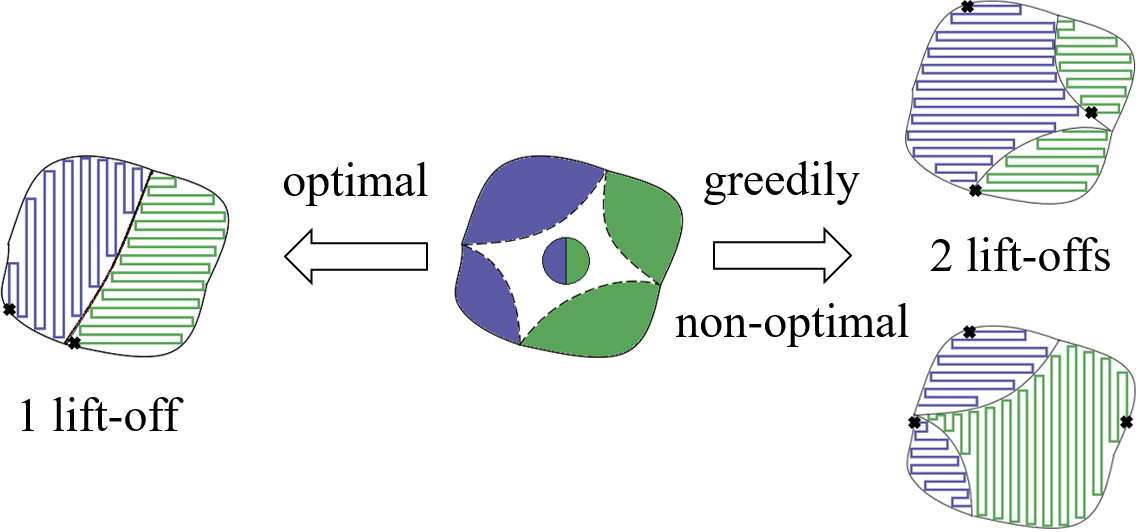}
\caption{
In this example, only the middle cell still needs solving. From the two available colours, whatever the choice to fill the cell in its entirety incurs an extra, uncessary lift-off. On the other hand, optimality is achieved when splitting the cell into two parts, 
with the two resulting sub-cells filled in with a different colors, requiring only $1$ lift-off. 
}\label{figsimpleexample}
\end{figure}

Finally, a topological graph is created, whose elements are cells. Each cell possesses an index, records the possible colors 
and the indices of its adjacet cells in order, as per the example in Fig.~\ref{figforcolor}. 
Since the number of colors is finite, the number of possible combinations of colors is also finite, which can be ordered as
$$\left\{
\begin{aligned}
&\{c_1\}, \cdots, \{c_n\}\\
&\{c_1, c_2\}, \cdots, \{c_1, c_n\}, \{c_2, c_3\}, \cdots, \{c_2, c_n\}, \cdots, \{c_{n-1}, c_n\}\\
&\cdots\ \mbox{(with all $i$-element combinations in the $i$-th row)}\\
&\{c_1, \cdots, c_n\}
\end{aligned}
\right.$$
For each combination of colors, the number of corresponding cells is finite, unless there are infinite many small cells with area zero, which is physically meaningless for the coverage task of the robots. In all, the number of cells must be finite. 

After creating the topological graph, the cellular decomposition process is transformed into painting all points in a graph with 
one of their available colors.
The number of 
solutions to ``painting'' the full graph means the number of coverage path segments, where discontinuities are required in between, with the minimum(s) as best solution. Two valid solutions exist for the example in Fig.~\ref{flowchart}. 
In summary, the proposed model of colouring a point in the surface to be covered means selecting a given IK solution for it, 
and the planning problem is thus transfered to designing a colour scheme for a topological configuration graph.

\section{Enumerative Solver for Cellular Decomposition}
\label{sectionenumerativesolver}
The difficulty of solving the coloring problem is that, although points are gathered into homogeneous cells, 
they can be filled in with different colors, instead of only being seen as a whole and drawn with a single colour. 
The counter-example in Fig.~\ref{figsimpleexample} illustrates this phenomenon. 
By efficiently discarding equivalent cellular decompositions, it can be proven that the total number of different cellular 
decompositions is finite, thus all optimal solutions are finitely solvable. 

\begin{figure}[t]
\centering
\subfigure[Example shows that it is sufficient to consider cutting path which starts and end at the topological edge endpoints.]{
\includegraphics[width=0.4\textwidth]{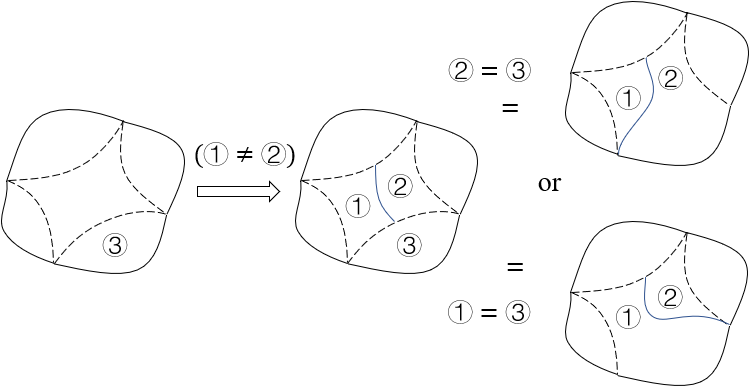}}
\subfigure[Example shows that is is unnecesary to consider cutting paths that stretch across edges. 
]{
\includegraphics[width=0.48\textwidth]{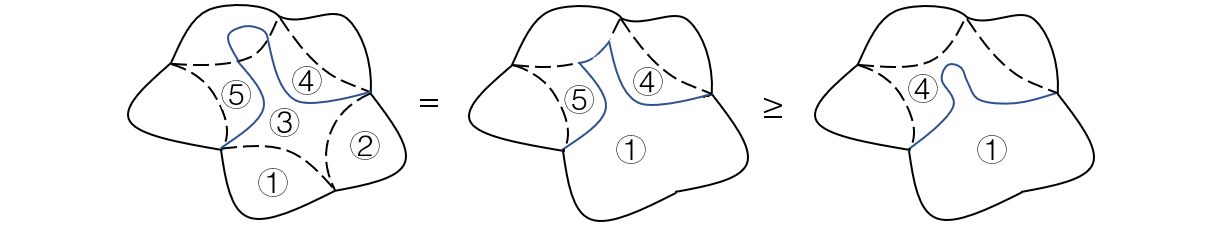}}
\subfigure[Example shows that intersecting cutting paths can be discarded.]{
\includegraphics[width = 0.48\textwidth]{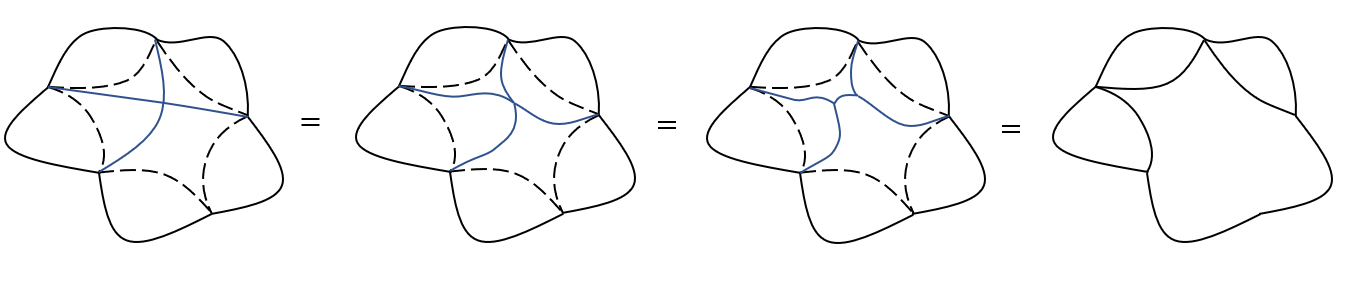}}
\caption{Unnecesary or sub-optimal cell divisions cases can be safely discarded.}
\label{figproof}
\end{figure}

\subsection{Finiteness of Divisions}
\label{subsectionproof}
Since any path starting and ending at the boundary of a cell will divide the cell into two parts, there are infinite many physical solutions of dividing a cell into parts. However, there are only finite classes of them from a topological structure viewpoint 
because of the equivalence of physical divisions in the number of lift-offs. 

Fig.~\ref{figproof}(a) shows how cutting paths which start or end at a point other than an endpoint on an edge are unnecessary and can be pruned. Let a cutting path end at an arbitrary point of the edge connecting with cell 3. From the definition of a cutting path, it implicitly enforces cell 1 and cell 2 having different colors.  
If $1\neq 3$ and $2\neq 3$ the division is trivial. 
Howewer, for the depicted cases when $1=3$ or $2=3$, any cutting paths that start at any endpoint of the edge are equivalent. Hence, for a complete solutoin it is sufficient to only consider cutting paths which start and end at the endpoint of and edge.  

Fig.~\ref{figproof}(b) shows how cutting paths which go across any edge are unnecessary. 
Let a cutting path go across an edge, then cell 4 and cell 5 are prevented from being colored together which leads to non-optimality, since they are separated physically by the cutting path.

Fig.~\ref{figproof}(c) shows that cutting paths need not go across each other. When two cutting paths intersect, 
the resulting cutting path segments can be continuously transformed back onto the existing topological edges, and can be safely diregarded.

In conclusion, only cutting paths which start and end at the endpoint of topological edges and do not go across each other need to be considered when considering options for cell subdivision, making the total number of topological divisions finite. 

\subsection{Solution to Simple Cells}
\label{sec:simple_cell}
\begin{figure}[t]
\centering
\includegraphics[width = 0.44\textwidth]{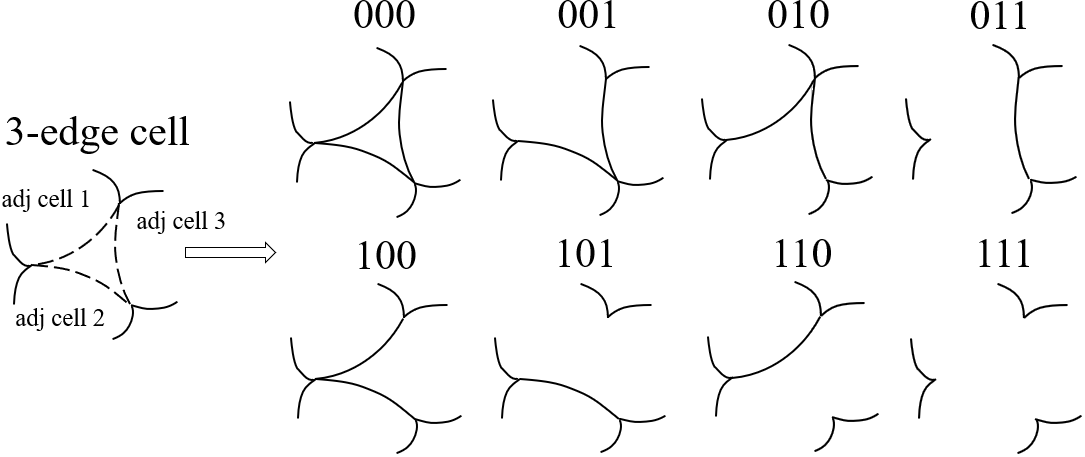}
\caption{Possible divisions of a 3-edge cell, as the most complex situation that need not further divisions. It can be observed how some of them are the same in terms of topological structure (e.g. $001, 010$ and $100$), 
or the division is impossible 
(e.g., $011$ is enforced, but cell 1 and 2 do not have the same colour), it is clearly already a finite problem, and the description of   further simplifications is omitted for clarity and space constraints.}
\label{figeasycell3}
\end{figure}

The following kinds of cells offer simple cases that can be solved directly without further divisions:
\begin{enumerate}
\item Cells containing less than four edges. They cannot be divided further into several cells with less number of topological edges. Fig.~\ref{figeasycell3} enumerates all possible topological divisions for a three-edge cell, which constitutes the most complicated case for direct enumeration. Binary number are used to represent the edge connectivity, $1$ (connected), $0$ (disconnected). It is thus easy to see that there are at most $8$ situations that require consideration.
\item Cells with only one possible colour.
\end{enumerate}

\subsection{Solution to Complex Cells}
\begin{figure}[t]
\centering
\includegraphics[width = 0.4\textwidth]{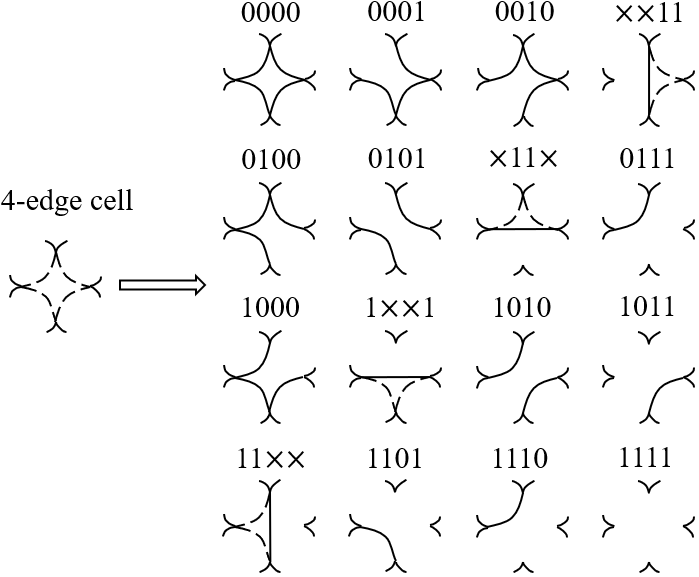}
\caption{The $2^4$ possible divisions of an 4-edge cell. For the cell which has more than 3 edges, the sub-cell may be created. In this figure, the connectivities that correspond to generating a sub-cell is $\times\times11, \times11\times, 1\times\times1, 11\times\times$. }\label{figeasycell4}
\end{figure}

The concept of using binary coding for a simple cell is extended to solve for an arbitrary $n$-edge cell, 
with the addition of ``$\times$'' for a yet unspecified connectivity state that emanates from these more complex scenarios. 
For this binary combination, there are less than $2^n\times m$ branches for an $n$-edge cell with $m$ possible colors, so the problem remains finite. 
An example of solving for $n=4$ is shown in Fig.~\ref{figeasycell4}.
In this case, a continuous connected state ($1$s) implies that part of this cell must be painted with the same colour as that of the adjacent cells. 
On the other hand, a connectivity of $0$ indicates that the topological edge between the cell and the corresponding adjacent cell is maintained, and as such the colors must be different. 

\begin{figure}[t]
\centering
\includegraphics[width = 0.3\textwidth]{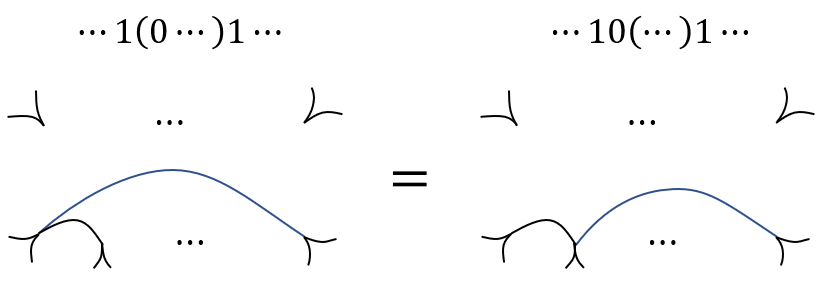}
\caption{The equivalence of moving the unconnected $0$ state outside the bracket. Following this, in order to reduce the number of edges of a sub-cell, the equivalent sub-cell connectivity arrangement on the right is enforced.}
\label{figconstraint}
\end{figure}

The unspecified state $\times$ indicates a sub-cell will be generated. 
A more detail discussion about the distinction between $0$ and $\times$ is warranted:
when connectivity cases such as the following arise 
$$1\times\times1, 1\times\times\times1, \cdots$$
the cell is divided into smaller parts whose colors are enforced to be different, i.e. the cell needs further subdivision.
Refer for instance to cases $\times\times11, \times11\times, 1\times\times1, 11\times\times$ in Fig.~\ref{figeasycell4}. 
The original cell transforms into a new one with fewer edges, since some (two for a 4-edge cell) 
edges are replaced by a single one.  
We use the bracket notation $(\cdots)$ in the binary number of the original cell to represent the generation of sub-cells.
In recursively applying the same division for the applicable sub-cells, any $n$-edge cell can thus be continuously divided into 
a set of cells with less edges, suitable to then be solved enumeratively as described. 
Since sub-cells are generated from an original one, there are extra constraints on its connectivity as specified by the previous division. 
As such, the following situation may arise:
\begin{enumerate}
\item Single $\times$ cannot form a sub-cell, because 
$$\cdots 1\times 1\cdots = \cdots 101\cdots \mbox{ or }\cdots 111\cdots$$
but both conditions of the right side are considered in other branches. This is why the $\times$ case does not apply 
to the 3-edge cell in Fig.~\ref{figeasycell3}.  
\item The $0$s can be freely moved outside the bracket, because of the equivalence 
$$\cdots\times1(0\times\cdots\times1)1\times\cdots = \cdots\times10(\times\cdots\times1)1\times\cdots$$ 
See Fig.~\ref{figconstraint}. 
The same is true for the right bracket based on the symmetry of the number list, since a cutting graph makes sense 
only when the inner two numbers at the boundary are $1$s. This is why there are no brackets in Fig.~\ref{figeasycell4}.
\item The new topological edge created by a cutting path must be retained as it is manually introduced as such 
(it will always be $0$, spliting a cell into different colours, it can not be be $1$ or $\times$).  
No extra possiblities appear after the division. 
\end{enumerate}

\section{Iterative Solver for Cellular Decomposition}
\label{sectioniterativesolver}

With the enumerative solver as the basic building block, the graph can now be solved iteratively.
Starting from a fully-unpainted graph, an unsolved cell is arbitrarily chosed and enumerativelly solved as 
described in Section~\ref{sectionenumerativesolver}. 
Assuming a the cell with $n$-edges with $m$ possible colors, there are \textit{at most} $2^n\times m$ possible divisions. A branch for each possible solution of the chosen cell is created, and in each branch the selected cell is filled in with the specfied colour (so it will not change any more). 
In the next iteration, an unsolved branch is selected, and an unsolved cell within it, and repeat the same steps.
 
Note that the constraints given by the solved adjacent cells significantly restrict the possible solutions, 
because the state of an edge resticts the connectivity of the cells on both sides. 
Through iterative execution, if there is a cell who cannot satisfy all constraints given by its solved adjacent cells, then the graph cannot be painted using the current state of the even partly-filled painting scheme, orelse a valid coloring scheme for the graph can be generated which uniquely specifies the configuration to polish each equivalent workspace point amongst the various valid IK solutions it may exhibit. 

The search algorithm runs on a deepest-first-searching (DFS) format, so that the memory requirements are reduced. 
As an exhaustive search protocol, all optimal physical cellular decompositions must be homeomorphic to one of the resulting schemes.

\subsection{Cost Calculation}
\begin{figure}[t]
\centering
\includegraphics[width = 0.4\textwidth]{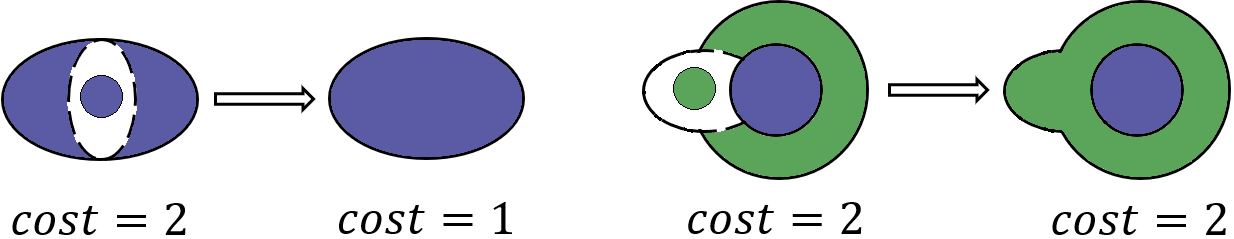}
\caption{Cost variation calculation. Left: the middle cell connects two distinct cells, so cost variation~(\ref{eq:cost}) is $1-2 = -1$. 
Right: two edges connect with the same adjacent cell, cost variation ~(\ref{eq:cost}) is $1-1 = 0$.} 
\label{figcost}
\end{figure}

The physical meaning of the cost for a (partly filled) graph is the number of colour segments in the current portion of the graph, 
i.e. 1 colour equals cost = 1.
The cost formulation is then described incrementally, whereby after a cell is solved,  
\begin{enumerate}
\item if its connectivity is all zero, then the cost will increase $1$ after coloring this cell, because this cell forms a new segment.
\item if its connectivity has only one $1$ connection to an already solved cell, then the cost will remain unchanged, since the cell colour can be filled homegeneously with the connected adjacent cell. 
\item if its connectivity has $i$ $1$s, there may exist multiple edges which connect the same adjacent cell, as per the illustration 
in Fig.~\ref{figcost}. In order to be consistent with the physical meaning of the cost, 
if these edges connect $j$ distinct solved cells, then the variation of cost is 
\begin{equation}
\Delta cost = 1-j
\label{eq:cost}
\end{equation}
\end{enumerate}

\section{Experimental Results}
\label{sectionexperiment}
The proposed algorithm performs non-repetitive coverage task using non-redundant manipulators of any dimension. 
To test it, simulation and experimental work have been implemented using a 5DoF manipulator to emulate a polishing 
task on an object's surface. For such endeavour, the final revolute joint of a typical 6 DoF manipulator is 
unnecessary given the rotating nature of the polishing tool itself, and real experiments are undertaken with a 
UR5 where the last link has been locked.

In the first simulated experiment, a hemispherical object is polished at different poses in the workplace, one arbitrarily set and the other being precisely designed to be fully reachable with the least number of lift-offs as a metric for evaluating the quality of the object placement to be inspected. 
The second simulated experiment shows how the proposed algorithm can directly influence the choice of configurations 
to avoid non-optimal configurations that invariably lead to unnecessary lift-offs altogether.
Finally, real-world experiments with a UR5 manipulator polishing a wok 
in free space and under the presence of obstacles proves the applicablity of the proposed algorithm.

In the results shown hereafter 
the environment contains the manipulator, the object being polished, and where applicable a ground plane and additional obstacles. 
Moreover, figures shown in this section are representative examples of arbitrary paths derived within the cells attained following the proposed optimal coverage solution.

\subsection{Covering a Hemispherical Object}
\begin{figure}[t]
\centering
\subfigure[]{\includegraphics[width = 0.10\textwidth]{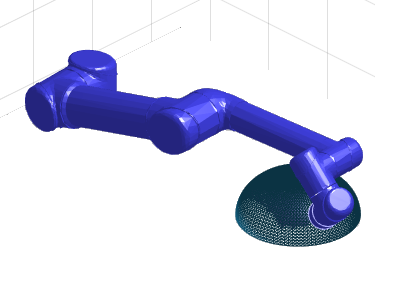}}
\subfigure[]{\includegraphics[width = 0.10\textwidth]{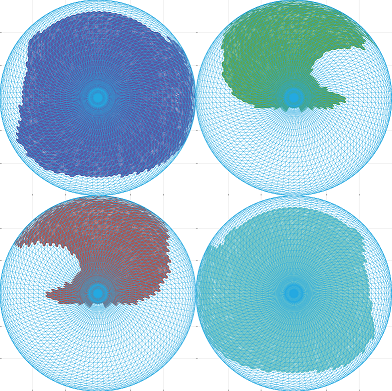}}
\subfigure[]{\includegraphics[width = 0.10\textwidth]{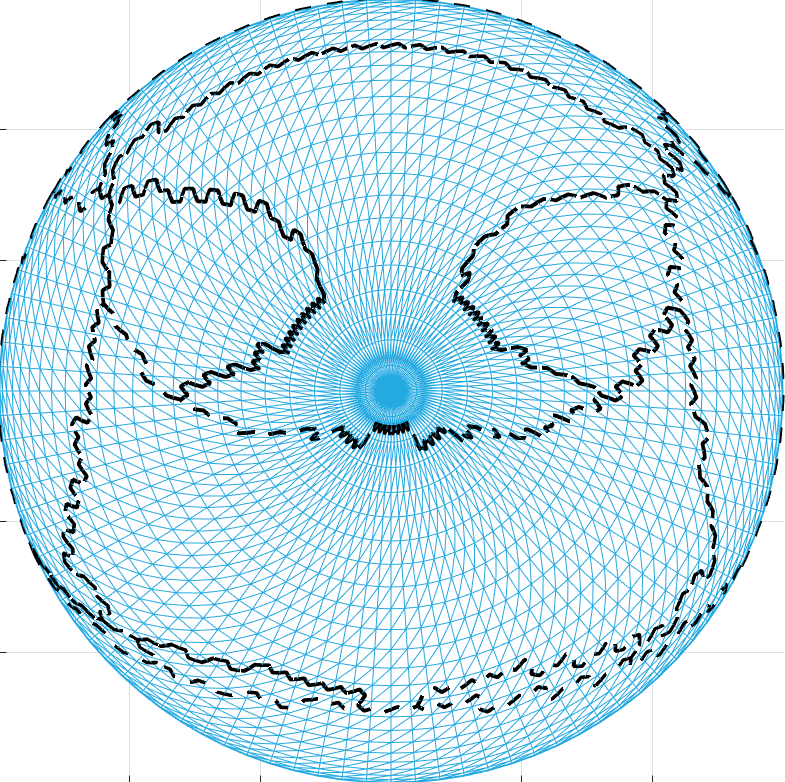}}
\subfigure[]{\includegraphics[width = 0.10\textwidth]{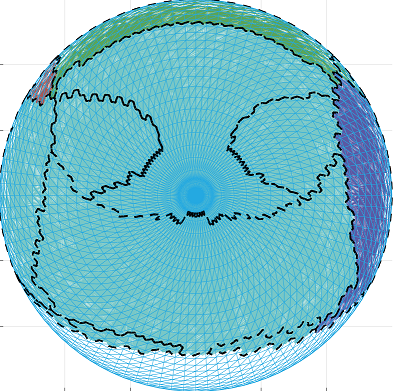}}
\caption{(a) Hemispherical object arbitrarily placed in the workspace. 
(b) Coloured cells of four valid configuration, chosen by the optimal solution shown in (d). 
(c) Topological graph. (d) One optimal solution requiring 3 lift-offs 
(note how the manipulator cannot fully cover the farthest part of the mesh - the bottom area in the optimal solution, unpainted). 
}\label{figflatwise}
\end{figure}

\begin{figure}[t]
\centering
\subfigure[]{\includegraphics[width = 0.1\textwidth]{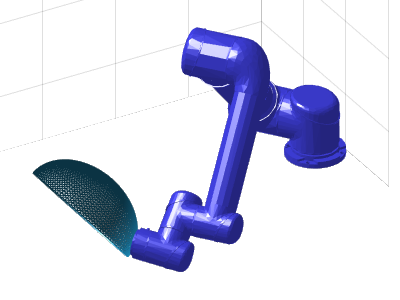}}
\subfigure[]{\includegraphics[width = 0.1\textwidth]{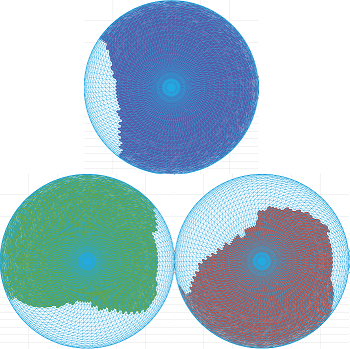}}
\subfigure[]{\includegraphics[width = 0.1\textwidth]{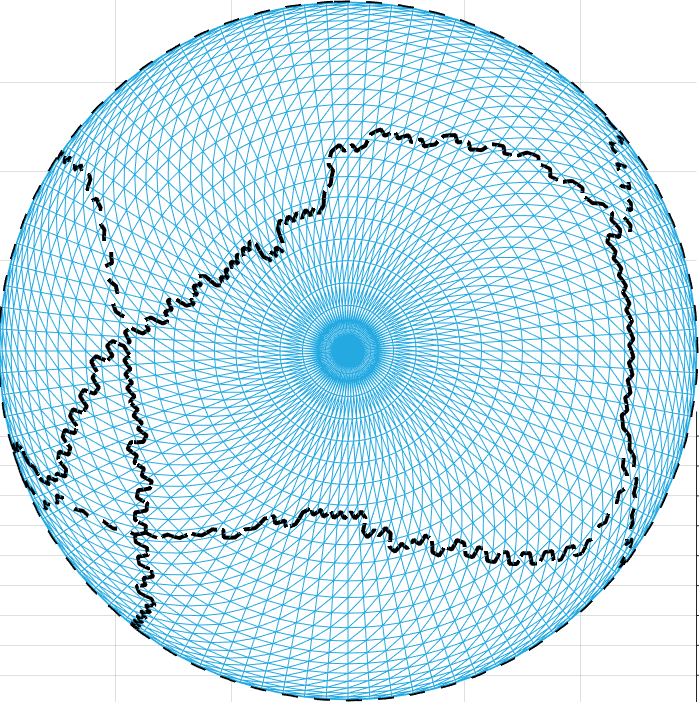}}
\subfigure[]{\includegraphics[width = 0.1\textwidth]{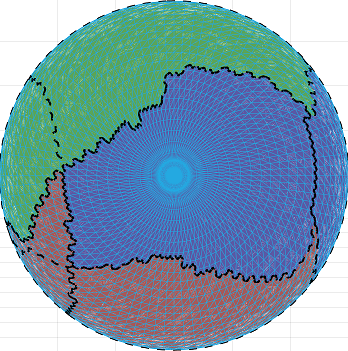}}
\caption{(a) Hemispherical object placed obliquely in the workspace to achieve full coverage. 
(b) Coloured cells of three valid configuration, chosen by the optimal solution shown in (d). 
(c) Topological graph. (d) One optimal solution requiring only 2 lift-offs and achieving fully coverage, with the object fully painted over). 
}\label{figsloped}
\end{figure}

A wok-like round mesh is used for this experiment. Results are collected in Fig.~\ref{figflatwise} and Fig.~\ref{figsloped}. 
In the former, the object is arbitrarily placed with respect to the robot, as would be the case, for instance, on an automated production line with unsorted objects are fed via a conveyor belt. A CPP is designed following the proposed scheme. With no criterion for the placement, the algorithm shows that such an object placement requires at least $3$ lift-offs to inspect the reachable area, yet fails in attaining full coverage (the farthest area, shown at the bottom of the mesh, is out-of-bounds).
Fig.~\ref{figsloped} illustrates the case where the proposed coverage strategy reveals a suitable pose for the object so that not only the required least number of lift-offs is decreased to $2$ when compared to the arbitrary placement, 
but the manipulator can fully cover the surface.

\subsection{Covering a Cylindrical Object}
Polishing of a half-pipe is employed in this simulation to demonstrate how the proposed algorithm can identify and bypass unnecessary configurations leading to ``traps'' that cause CPPs with extra lift-offs. 

\begin{figure}[t]
\centering
\subfigure[]{\includegraphics[width = 0.13\textwidth]{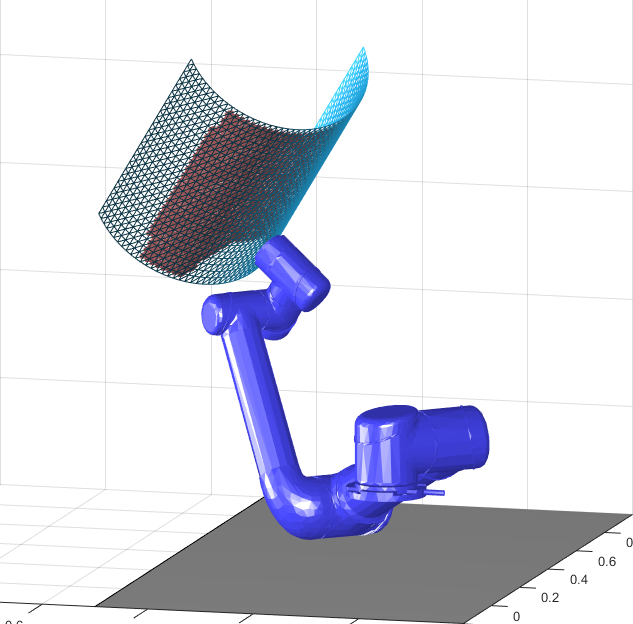}}
\subfigure[]{\includegraphics[width = 0.13\textwidth]{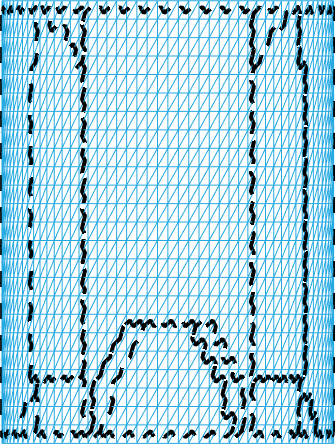}}
\subfigure[]{\includegraphics[width = 0.13\textwidth]{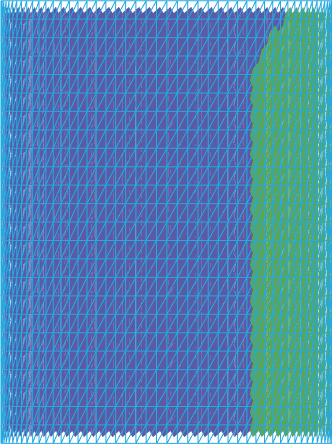}}
\caption{(a) A cylindrical half-pipe object. (b) The topological graph. (c) One optimal solution which requires only 1 lift-off.}\label{fighalfpipe}
\end{figure}

\begin{figure}[htb]
\centering
\includegraphics[width = 0.4\textwidth]{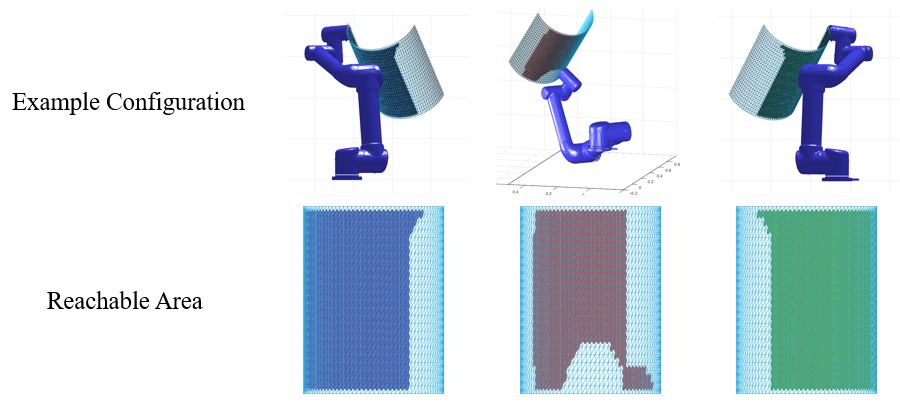}
\caption{Example with three different configurations and their reachable area. Once any configuration belonging to the middle cell is chosen to cover the surface, sooner or later it will have to change to the left and right cells to finish the full coverage of the surface, thus incurring a wasteful lift-off.}
\label{figthreeexamplepose}
\end{figure}

The pipe is placed obliquely to achieve full coverage. Surface normals vary along the arc length of the cylinder 
over $\pi$ radians, which causes increased difficulty in kinematic terms for the manipulator 
to sustain the desidered polishing operation. 
The topological graph and optimal solution are shown in Fig.~\ref{fighalfpipe}, where it can be seen that the full coverage task requires only $1$ lift-off. 
However, there are many valid configurations which lead to non-optimality. 
See Fig.~\ref{figthreeexamplepose} for an example of such ``trap'' configurations. 
The configurations on the left and right are the ones finally chosen by one of the optimal solutions shown in Fig.~\ref{fighalfpipe}. 
However, while the configuration in the middle can cover a large area without lift-offs, and would therefore be equally likely to be chosen if the IK solutions were to be selected randomly or in a greedy fashion, it cannot reach the corners of the mesh (eventually covered by the other two configurations). Hence, should any configuration from the middle be selected to trace 
the object, after coverage of (a section) of the middle part of the pipe, sooner or later the CPP will have to undertake one unnecesary lift-off for full coverage inspection, leading to non-optimality when compared to the case shown in Fig.~\ref{fighalfpipe}. The proposed algorithm will provide all optimal solutions, with none of them using the middle colored cell.

\subsection{Real World Experiments in the Presence of Obstacles}
A Universal Robots UR5 manipulator is employed in real experiments to polish the outer surface of a wok to show a physical coverage path generated based on the proposed cellular decomposition method. 
The actual physical coverage path uses a simple back and fore motion within the resulting cells, whilst the lift-off concatenation between paths segments belonging to different cells are manually demonstrated. 
The manipulator ioperates as a 5DOF. 
Since hybrid position/force control~\cite{solanes2019robust} is beyond the scope of this work, contact is restricted to position control. 

\begin{figure}[t]
\centering
\subfigure[]{\includegraphics[width = 0.14\textwidth]{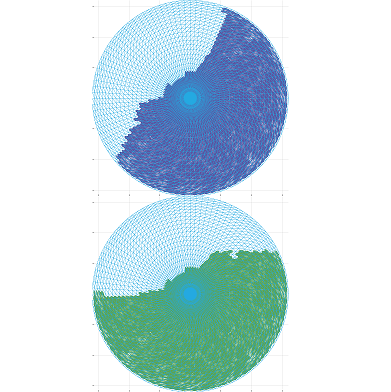}}
\subfigure[]{\includegraphics[width = 0.14\textwidth]{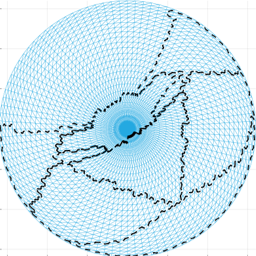}}
\subfigure[]{\includegraphics[width = 0.14\textwidth]{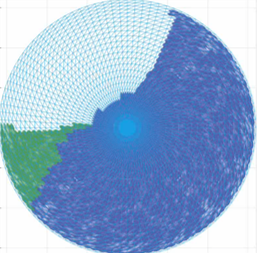}}
\subfigure[]{\includegraphics[width = 0.2\textwidth]{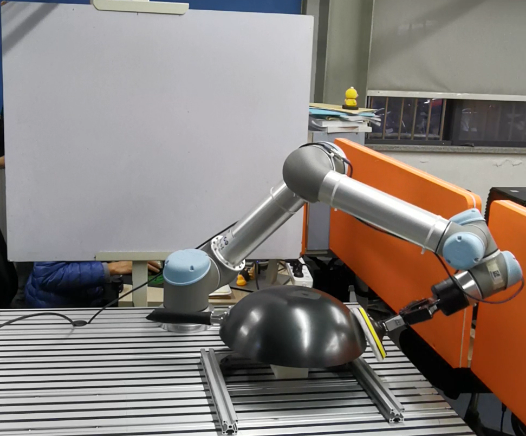}\label{fig:real_free_extreme1}}
\subfigure[]{\includegraphics[width = 0.2\textwidth]{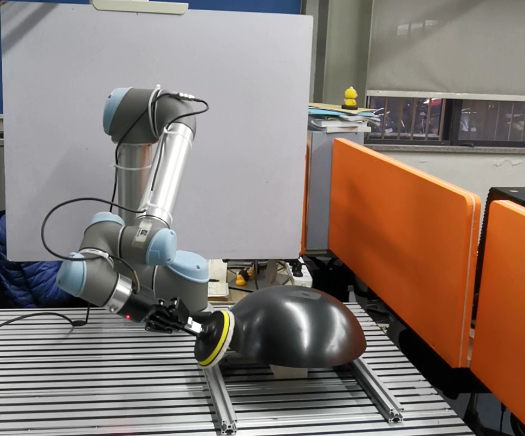}\label{fig:real_free_extreme2}}
\caption{(a) Reachable coloured cells of two valid configurations, chosen by the optimal solution shown in (c). 
(b) Topological graph. (c) One of the optimal solutions requiring only 1 lift-off. 
The cutting path is arbitrary. (d),(e) Examples of extreme poses in the two types of configurations.}
\label{fig_realworld_no}
\end{figure}

Fig.~\ref{fig_realworld_no} illustrates the results. Given the location of the wok, it can be seen how the nearest 
part of the wok is unreachable. 
As can be observed in Figs.~\ref{fig:real_free_extreme1} and~\ref{fig:real_free_extreme1}, 
the manipulator must keep its wrist configured in the ``above'' the fore-arm configuration in order to 
avoid collisions, which leads to the two shoulder-left and shoulder-right configuration solutions. 
The total number of lift-offs is $1$. 
Note that any division keeping the resulting cell connectivity guarantees full (reachable) coverage and is optimal in the minimum number of lift-offs, so the cutting path dividing the final cell is arbitrary. 

\begin{figure}[t]
\centering
\subfigure[]{\includegraphics[width = 0.14\textwidth]{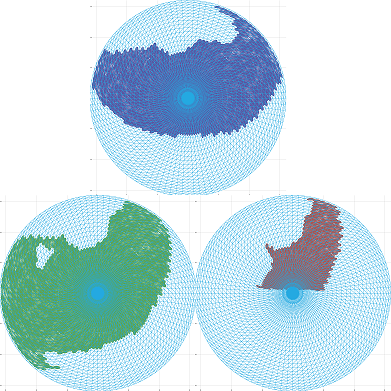}}
\subfigure[]{\includegraphics[width = 0.14\textwidth]{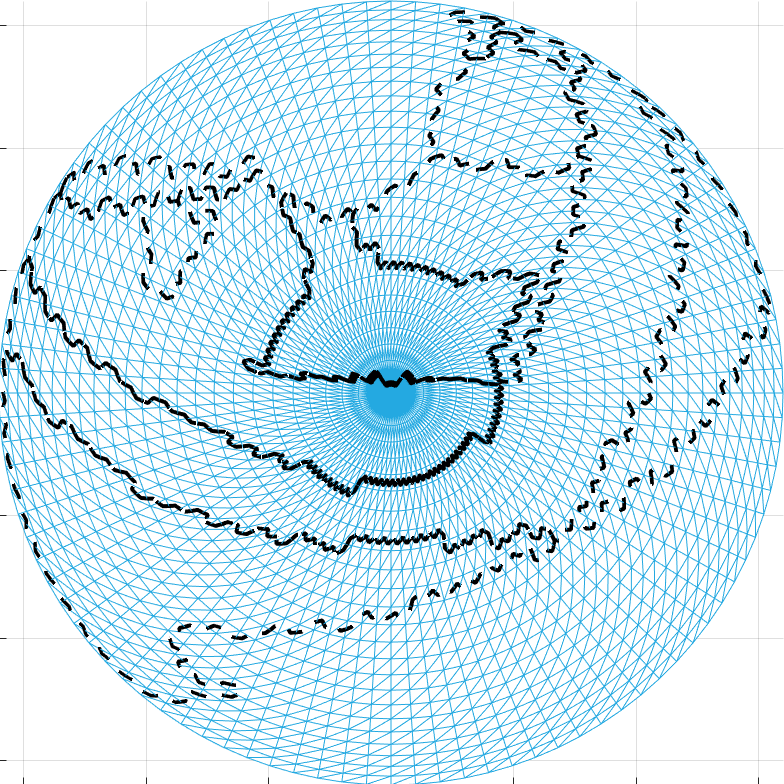}}
\subfigure[]{\includegraphics[width = 0.14\textwidth]{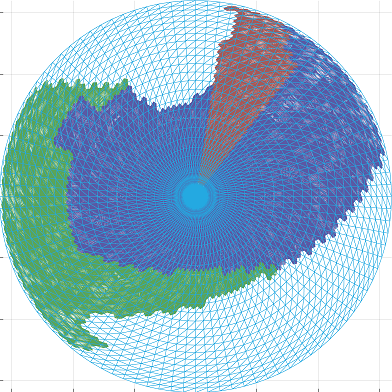}}
\subfigure[]{\includegraphics[width = 0.14\textwidth]{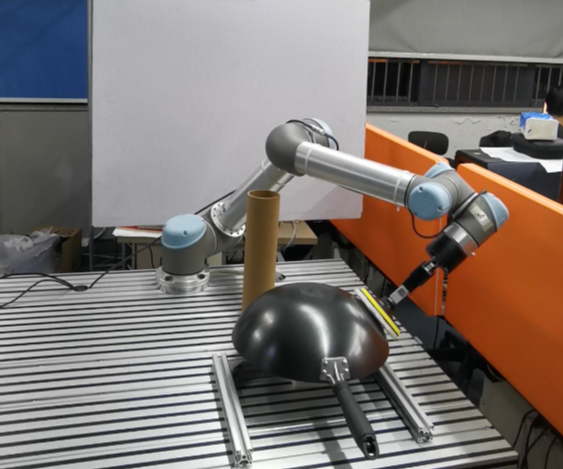}}
\subfigure[]{\includegraphics[width = 0.14\textwidth]{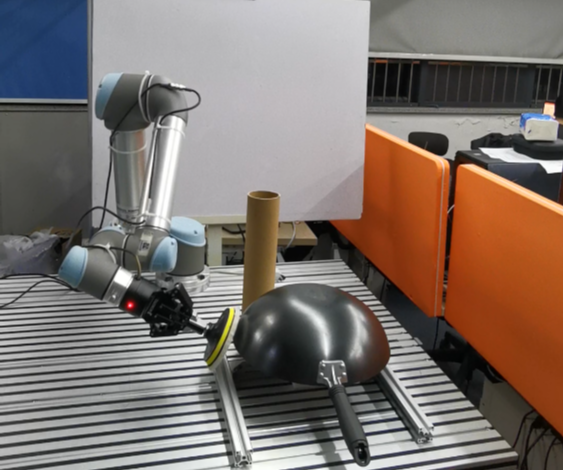}}
\subfigure[]{\includegraphics[width = 0.14\textwidth]{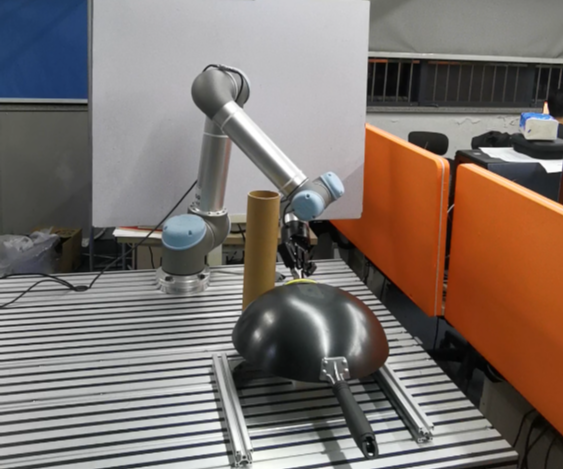}}
\caption{(a) Reachable coloured cells of three valid configuration, chosed by the solution in (c). 
(b) Topological graph. (c) One of the optimal solutions, requiring 2 lift-offs. 
(d),(e),(f) Example of the three kinds of configurations, where
(d) Example of the shoulder-left configuration adopted to avoid collision between the upper-arm and the obstacle. 
(e) Example of the wrist above the fore-arm configuration, so that the wrist avoids colliding with the fore-arm. 
(f) An example of the only valid configuration to cover the points in the brown cell, situating the wrist below the fore-arm.}\label{fig_realworld_with}
\end{figure}

A more interesting example arises when the motion of the manipulator is obstructed by the cylindrical obstacle 
depicted in Fig.~\ref{fig_realworld_with}. Since the obstacle may collide with the upper-arm, fore-arm or the EE, 
and the wrist may collide with the fore-arm, 
the resulting colour cell decomposition and the topological graphs is more complex. As such, to avoid collisions, 
the least number of lift-offs is shown to be 2. 

\section{Conclusion}
\label{sectionconclusion}

A novel proposition for the coverage planning problem has been develop in this work. The focus is set on the minimum
number of coverage path discontinuities as key metric, predicated on the need for manipulator tasks such as polishing, painting or 
deburring to curtail the number of lift-offs for proficient results.

To achive this goal, instead of considering the design of a coverage path in the traditional sense, this research considers the global optimal cellular decomposition problem in joint-space to assemble joint-space partitions with minimum sets. 
In noting that IK mapping from the reachable points in the workspace to a single set of configurations is injective, 
colouring a point in the surface to be covered means selecting a given IK solution for it, and the planning problem is transfered to designing a colour scheme for a topological configuration graph. In that way, the key concern is joint-space continuity of the cells. 

The proposed scheme thus provides two relevant conributions to the CPP problem in relation to optimal discontinuities:  
Proof that the least number of discontinuities, or ``lift-offs'', is independent of the choice of coverage path.  And,
by sugesting a novel cellular decomposition strategy for the efficient discarding of equivalent cells, this work proves 
that the total number of different cellular decompositions is finite, thus all optimal solutions are finitely solvable. 

After applying any conventional CPP algorithm in each resulting cell, the nominated algorithm thus generates a coverage path containing the least number of discontinuities. As a direct corollary, the result strategy can be applied to the result of other cellular decomposition methods in the literature (e.g. Morse-based), to produce the least number of discontinuities obeying the given cellular decomposition method. It can also be exploited as a criterion to evaluate the placement quality of a manipulator or object in the workspace for minimal lift-off coverage paths. A systematic way to resolve this issue is left for future work.
Extensive simulation and real-world implementation results on a 5 DoF manipulator are presented, suplemented by a video, to prove the validity of the proposed strategy in challenging realistic conditions.

\section*{Acknowledgment}
This work was supported in part by the National Key Research and Development Program of China under Grant 2018YFB1309300 and in part by the National Nature Science Foundation of China under Grant U1609210 and Grant 61903332. 

\bibliographystyle{ieeetr} 
\bibliography{min_removal}

\begin{thebibliography}{10}

\bibitem{choset2001coverage}
H.~Choset, ``Coverage for robotics – a survey of recent results,'' {\em
  Annals of Mathematics and Artificial Intelligence}, vol.~31, no.~1,
  pp.~113--126, 2001.

\bibitem{galceran2013a}
E.~Galceran and M.~Carreras, ``A survey on coverage path planning for
  robotics,'' {\em Robotics and Autonomous Systems}, vol.~61, no.~12,
  pp.~1258--1276, 2013.

\bibitem{Oriolo2005Motion}
G.~Oriolo and C.~Mongillo, ``Motion planning for mobile manipulators along
  given end-effector paths,'' in {\em Robotics and Automation, 2005. ICRA 2005.
  Proceedings of the 2005 IEEE International Conference on}, 2005.

\bibitem{porta2010path}
J.~Porta and L.~Jaillet, ``Path planning on manifolds using randomized
  higher-dimensional continuation,'' vol.~68, pp.~337--353, 01 2010.

\bibitem{Porta2012Randomized}
J.~M. Porta, L.~Jaillet, and O.~Bohigas, ``Randomized path planning on
  manifolds based on higher-dimensional continuation,'' {\em International
  Journal of Robotics Research}, vol.~31, no.~2, pp.~201--215, 2012.

\bibitem{hassan2018a}
M.~Hassan and D.~Liu, ``A deformable spiral based algorithm to smooth coverage
  path planning for marine growth removal,'' pp.~1913--1918, 2018.

\bibitem{cheah2003brief}
C.~C. Cheah, S.~Kawamura, and S.~Arimoto, ``Brief stability of hybrid position
  and force control for robotic manipulator with kinematics and dynamics
  uncertainties,'' {\em Automatica}, vol.~39, no.~5, pp.~847--855, 2003.

\bibitem{heck2015switched}
D.~Heck, A.~Saccon, N.~Wouw, and H.~Nijmeijer, ``Switched position-force
  tracking control of a manipulator interacting with a stiff environment,''
  {\em Proceedings of the American Control Conference}, vol.~2015,
  pp.~4832--4837, 07 2015.

\bibitem{mirrazavi2018a}
S.~S.~S. Mirrazavi and B.~Aude, ``A dynamical-system-based approach for
  controlling robotic manipulators during noncontact/contact transitions,''
  {\em IEEE Robotics \& Automation Letters}, vol.~3, no.~4, pp.~2738--2745.

\bibitem{solanes2018adaptive}
J.~E. Solanes, L.~Gracia, P.~Munozbenavent, A.~Esparza, J.~V. Miro, and
  J.~Tornero, ``Adaptive robust control and admittance control for
  contact-driven robotic surface conditioning,'' {\em Robotics and
  Computer-integrated Manufacturing}, vol.~54, pp.~115--132, 2018.

\bibitem{solanes2019robust}
J.~E. Solanes, L.~Gracia, P.~Munozbenavent, J.~V. Miro, C.~Perezvidal, and
  J.~Tornero, ``Robust hybrid position-force control for robotic surface
  polishing,'' {\em Journal of Manufacturing Science and
  Engineering-transactions of The Asme}, vol.~141, no.~1, p.~011013, 2019.

\bibitem{lumelsky1990dynamic}
V.~J. {Lumelsky}, S.~{Mukhopadhyay}, and K.~{Sun}, ``Dynamic path planning in
  sensor-based terrain acquisition,'' {\em IEEE Transactions on Robotics and
  Automation}, vol.~6, no.~4, pp.~462--472, 1990.

\bibitem{choset2000exact}
H.~Choset, E.~U. Acar, A.~A. Rizzi, and J.~Luntz, ``Exact cellular
  decompositions in terms of critical points of morse functions,'' vol.~3,
  pp.~2270--2277, 2000.

\bibitem{Acar2002Morse}
E.~U. Acar, H.~Choset, A.~A. Rizzi, P.~N. Atkar, and D.~Hull, ``Morse
  decompositions for coverage tasks,'' {\em The International Journal of
  Robotics Research}, vol.~21, no.~4, pp.~331--344, 2002.

\bibitem{choset2005principles}
H.~Choset, K.~M. Lynch, S.~Hutchinson, G.~Kantor, W.~Burgard, L.~Kavraki, and
  S.~Thrun, {\em Principles of Robot Motion: Theory, Algorithms, and
  Implementation}.
\newblock MIT Press, 2005.

\bibitem{choset1998coverage}
H.~Choset and P.~Pignon, ``Coverage path planning: The boustrophedon cellular
  decomposition,'' pp.~203--209, 1998.

\bibitem{choset2000coverage}
H.~Choset, ``Coverage of known spaces: The boustrophedon cellular
  decomposition,'' {\em Autonomous Robots}, vol.~9, no.~3, pp.~247--253, 2000.

\bibitem{choset2000sensor-based}
H.~Choset and J.~W. Burdick, ``Sensor-based exploration: The hierarchical
  generalized voronoi graph,'' {\em The International Journal of Robotics
  Research}, vol.~19, no.~2, pp.~96--125, 2000.

\bibitem{Atkar2003Towards}
P.~Atkar, H.~Choset, and A.~Rizzi, ``Towards optimal coverage of 2-dimensional
  surfaces embedded in $\mathbb{R}^3$: choice of start curve,'' in {\em
  Proceedings of 2003 IEEE/RSJ International Conference on Intelligent Robots
  and Systems (IROS '03)}, vol.~4, pp.~3581 -- 3587, October 2003.

\bibitem{huang2001optimal}
W.~H. {Huang}, ``Optimal line-sweep-based decompositions for coverage
  algorithms,'' in {\em Proceedings 2001 ICRA. IEEE International Conference on
  Robotics and Automation (Cat. No.01CH37164)}, vol.~1, pp.~27--32 vol.1, May
  2001.

\bibitem{jimenez2007optimal}
P.~A. Jimenez, B.~Shirinzadeh, A.~E. Nicholson, and G.~Alici, ``Optimal area
  covering using genetic algorithms,'' {\em international conference on
  advanced intelligent mechatronics}, pp.~1--5, 2007.

\bibitem{paus2017a}
F.~Paus, P.~Kaiser, N.~Vahrenkamp, and T.~Asfour, ``A combined approach for
  robot placement and coverage path planning for mobile manipulation,'' in {\em
  2017 IEEE/RSJ International Conference on Intelligent Robots and Systems
  (IROS)}, 2017.

\bibitem{yoshikawa1990translational}
T.~Yoshikawa, ``Translational and rotational manipulability of robotic
  manipulators,'' {\em American Control Conference}, vol.~27, pp.~228--233,
  1990.

\end{thebibliography}

\ifCLASSOPTIONcaptionsoff
  \newpage
\fi

\begin{IEEEbiography}[{\includegraphics[width=1.1 in,height=1.25 in,clip,keepaspectratio]{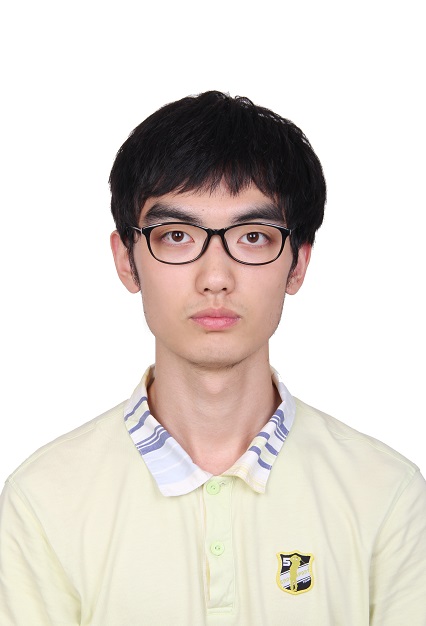}}]{Tong Yang} 
is currently pursuing the Ph.D. degree with the Institute of Cyber-Systems and Control, Department of Control Science and Engineering, Zhejiang University, Hangzhou, China. His current research interests include robot planning and control.
\end{IEEEbiography}

\begin{IEEEbiography}[{\includegraphics[width=1.1 in,height=1.25 in,clip,keepaspectratio]{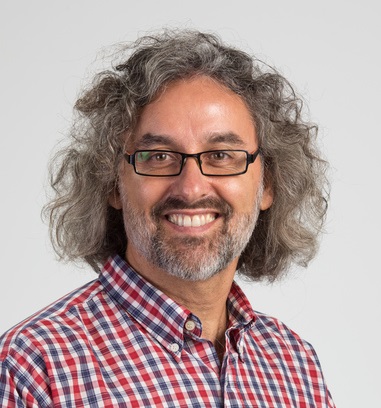}}]
{Jaime Valls Miro} received his B.Eng. and M.Eng. in Computer Science (Systems Engineering) from the Valencia Polytechnic University (UPV, Spain), in 1990 and 1993 respectively. He received his Ph.D. in robotics and control systems from Middlesex University (UK) in 1998, and worked in the underwater robotics industry as a software and control systems analyst until 2003. In 2004 he joined the Centre for Autonomous Systems in UTS (Australia), where he is currently an Associate Professor. His areas of interest span across the field of robotics, most notably modelling sensor behaviours for perception and mapping, computational Intelligence in HRI , and robot navigation. 
\end{IEEEbiography}


\begin{IEEEbiography}[{\includegraphics[width=1.1 in,height=1.25 in,clip,keepaspectratio]{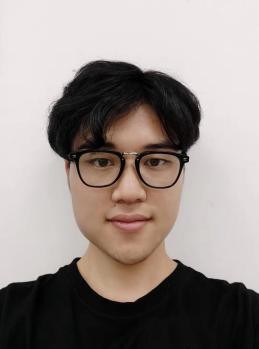}}]{Qianen Lai}
is currently pursuing the master degree with the Institute of Cyber-Systems and Control, Department of Control Science and Engineering, Zhejiang University, Hangzhou, China. His current research interests include robot grasping and robotics vision.
\end{IEEEbiography}
\begin{IEEEbiography}[{\includegraphics[width=1.1 in,height=1.25 in,clip,keepaspectratio]{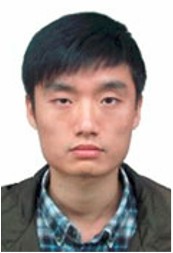}}]{Yue Wang}
 is currently a Post-Doctotral Fellow with the Institute of Cyber-Systems and Control, Department of Control Science and Engineering, Zhejiang University, Hangzhou, China. His current research interests include mobile robotics and robot perception.
\end{IEEEbiography}

\begin{IEEEbiography}[{\includegraphics[width=1.1 in,height=1.25 in,clip,keepaspectratio]{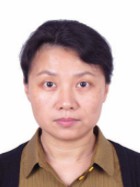}}]{Rong Xiong}
is currently a Professor with the Institute of Cyber-Systems and Control, Department of Control Science and Engineering, Zhejiang University, Hangzhou, China. Her current research interests include robot planning and robotics vision.
\end{IEEEbiography}




\end{document}